# Recurrent Neural Network Based Hybrid Model of Gene Regulatory Network

Khalid Raza[1*] and Mansaf Alam[1]

[1]Department of Computer Science, Jamia Millia Islamia (Central University), New Delhi-110025, India.

*kraza@jmi.ac.in

**Abstract**

Systems biology is an emerging interdisciplinary area of research that focuses on study of complex interactions in a biological system, such as gene regulatory networks. The discovery of gene regulatory networks leads to a wide range of applications, such as pathways related to a disease that can unveil in what way the disease acts and provide novel tentative drug targets. In addition, the development of biological models from discovered networks or pathways can help to predict the responses to disease and can be much useful for the novel drug development and treatments. The inference of regulatory networks from biological data is still in its infancy stage. This paper proposes a recurrent neural network (RNN) based gene regulatory network (GRN) model hybridized with generalized extended Kalman filter for weight update in backpropagation through time training algorithm. The RNN is a complex neural network that gives a better settlement between the biological closeness and mathematical flexibility to model GRN. The RNN is able to capture complex, non-linear and dynamic relationship among variables. Gene expression data are inherently noisy and Kalman filter performs well for estimation even in noisy data. Hence, non-linear version of Kalman filter, i.e., generalized extended Kalman filter has been applied for weight update during network training. The developed model has been applied on DNA SOS repair network, IRMA network, and two synthetic networks from DREAM Challenge. We compared our results with other state-of-the-art techniques that show superiority of our model. Further, 5% Gaussian noise has been added in the dataset and result of the proposed model shows negligible effect of noise on the results.

## 1. Introduction

Due to rapid advancement in high-throughput techniques for the measurement of biological data the attention of the research community has shifted from a reductionist view to a more complex understanding of biological system. The enriched understanding about genomes of various organisms, together with advancement in microarray technology, has provided researchers enormous opportunities for the development of computational and mathematical model of biological networks. The major objectives for modeling biological networks are (i) to present a synthetic view of the available biological knowledge in the form of network to better understand interactions and relationships on a holistic level, (ii) allow researchers to make predictions about gene function that can then be tested at the bench (iii) allow study of network's dynamical behavior (iv) complexity of molecular and cellular interactions requires modeling tools that can be used to design and interpret biological experiments, (v) it is essential for understanding the cell behavior, which in-turn leads to better diagnosis, (vi) predicts interactions between biological macromolecules not known so far, and (vii) allows for drug effect simulation.

The discovery of biological pathways or regulatory networks leads to a wide range of applications, such as pathways related to a disease can unveil in what way the disease acts and provide novel tentative drug targets. The development of biological models from discovered networks or pathways can also help to predict the responses to disease and can be much useful for the novel drug development and treatments. The high-throughput measurement techniques and plenty of data produced have brought the hope that we would be able to discover entire regulatory networks from these data. Unfortunately, the data and the biological systems that produce the data are often noisy and also biological processes are not well understood. These are some of the challenges that must be tackled. Hence, there is need of such kind of computational approach that may be able to tolerate noises in the data without affecting much in results.

The gene regulatory networks (GRNs) are a kind of "biological circuit" that may consists of input signaling pathways, regulatory proteins that integrate the input signals, target genes, and the RNA and proteins produced from these target genes. These networks may also include dynamic feedback loops that can further regulate gene expression. The GRNs are the systematic biological networks that describe regulatory interactions among genes in the form of a graph, where node represents genes and edges their regulatory interactions. A directed edge runs from one node to another. If there is transfer of information in both the directions, then the edge is undirected. Mathematically, a GRN is mapped to a graph $G = \{V, E\}$, where vertices $V$ represents the set of $N$ genes $\{G_1, G_2, \ldots, G_N\}$ and edges $E$ represents the regulatory interactions between genes in $V$. GRNs have been proved to be a very useful tool used to describe the complex dependencies between key developmental transcription factors (TFs), their target genes and regulators (Lee & Yang, 2008). In this work, we proposed a recurrent neural network (RNN) based hybrid model of GRN where generalized extended Kalman filter has been hybridized with training algorithm for weight update function. First, we briefly described RNN, RNN as GRN model and its working, Kalman filter and generalized extended Kalman filter for weight update in backpropagation through time training algorithm. The proposed hybrid model has been tested on four benchmark datasets. Section 2 describes some state-of-the-art techniques for modeling GRN. In section 3, recurrent neural network, training algorithm, Kalman filter and proposed model are described. Section 4 presents results and discussion and finally Section 5 concludes the paper.

## 2.    Modeling Gene Regulatory Networks: State-of-the-art Techniques

Today, one of the most exciting problems in systems biology research is to decipher how the genome controls the development of complex biological system. The GRNs help in identifying the interactions between genes and provide fruitful information about the functional role of individual genes in a cellular system. They also help in diagnosing various diseases including cancer. It indicates the assembly of regulatory effects and interactions between genes in a biological system and reveals complex interactions among genes under various stimuli or environmental conditions. Thus, modeling of GRNs are required that enables to decode the gene interaction mechanism in various kind of stimuli and further these information can be utilized for the prediction of the adverse effects of new drugs (Bower & Bolouri, 2004). In the last several decades, many computational methods have been proposed to model, analyze and infer complex regulatory interactions. These techniques are Directed graph, Boolean networks (Kauffman, 1969; Liang et al., 1998; Akutsu et al., 1999; Silvescu & Honavar, 2001; Shmulevich et al., 2002; Yun & Keong, 2004; Martin et al., 2007; Shmulevich & Dougherty, 2010), Bayesian networks (Friedman et al., 2000; Husmeier, 2003), Petri nets (Koch et al., 2005; Remy et al., 2006), linear and non-linear ordinary differential equations (ODEs) (Chen et al., 1999; Tyson et al., 2002; Jong & Page, 2008), machine learning approaches (Weaver et al., 1999; Kim et al., 2000; Vohradsky, 2001; Keedwell et al., 2002; Huang et al., 2003; Tian & Burrage, 2003; Zhou et al., 2004; Xu et al., 2004; Hu et al., 2005; Jung & Cho, 2007, Xu et al., 2007a; Xu et al., 2007b; Chiang & Chao, 2007; Lee & Yang, 2008; Datta et al., 2009; Zhang et al., 2009; Maraziotis et al., 2010; Ghazikhani et al., 2011; Liu et al., 2011; Kentzoglanakis, 2012; Noman et al., 2013), etc. For review of the modeling techniques and the subject, refer to (Jong, 2002; Wei et al., 2004; Schlitt & Brazma, 2007; Cho et al., 2007; Karlebach & Shamir, 2008; Swain et al., 2010; Sirbu et al., 2010; Mitra et al., 2011; Raza & Parveen, 2012, Raza & Parveen, 2013).

Most popularly known and widely used machine learning technique - artificial neural network (ANN) is a computational model, inspired by structural and functional aspects of biological nervous systems. The capabilities of ANNs to learn from the data, approximate any multivariate nonlinear function and its robustness to noisy data make ANN a suitable candidate for modeling gene regulatory interactions from gene expression data. Several variants of ANNs have been successfully applied, for modeling gene regulatory interactions, including multilayer perceptrons (Kim et al., 2000; Huang et al., 2003; Zhou et al., 2004), self-organizing maps (SOM) (Weaver et al., 1999) and recurrent neural networks (RNNs) (Vohradsky, 2001; Keedwell et al., 2002; Tian & Burrage, 2003; Xu et al., 2004; Hu et al., 2005; Chiang & Chao, 2007; Xu et al, 2007a; Xu et.al, 2007b; Ghazikhani et al., 2011; Noman et al., 2013; Raza et al., 2014).

Vohradsky (Vohradsky, 2001) proposed an ANN based model of gene regulation assuming that the regulation effect on gene expression of a particular gene can be expressed in the form of ANN. Each neuron in the network represents a particular gene and the connection between the neurons represents regulatory interactions. Here each layer of the ANN represents the level of gene expression at time *t* and output of a neuron at time *t+Δt* can be derived from these expressions. The advantage of this model is that it is continuous, uses a transfer function to transform the inputs to a shape close to those observed in natural processes and does not use artificial elements. The limitation is that it consists of several parameters that must be computed. Keedwell and his collaborators (Keedwell et al., 2002) also applied ANN in the purest form for the reconstruction of GRNs from microarray data. Architecture of the neural network was quite simple when dealing with Boolean networks and standard feed-forward backpropagation learning method has been applied.

Stochastic neural network model in the framework of a coarse-grained approach was proposed by Tian and Burrage (Tian & Burrage, 2003) for better description of the GRNs. The model is able to represent both intermediate regulation as well as chance events in gene expression. Poisson random variables are applied to represent chance events. Xu et al., (2004) applied RNN to infer GRN. An evolutionary algorithm based techniques, called particle swarm optimization (PSO), and most commonly used training algorithm backpropagation through time (BPTT) has been applied. The PSO is an evolutionary technique that tries to optimize a solution by iteratively improving a candidate solution. Hu et al. (Hu et al., 2005) has proposed a general recurrent neural network (RNN) model for the reverse-engineering of GRNs and to learn their parameters. RNN has been deployed due to its capability to deal with complex temporal behaviour of genetic networks. In this model, time delay between the output of a gene and its effect on another gene has been incorporated.

Chiang and Chao (Chiang & Chao, 2007) introduced a genetic algorithm (GA) hybridized with RNN (GA-RNN) for finding feed-forward regulated genes. This GA-RNN hybrid method constructs various kinds of regulatory modules. RNN controls the feed-forward and feed-backward loop in regulatory module and GA provide a way to search for commonly regulated genes. Xu and colleagues (Xu et al, 2007b) proposed a hybridized form of PSO and RNN, called PSO-RNN. In this approach, gene interaction is demonstrated by a connection weight matrix and PSO-based searching algorithm is presented to uncover genetic interactions that best fit the time series data and analyse possible genetic interactions. PSO is applied to train the network and find out network parameters. Another RNN-based model with three different training algorithms, differential evolution (DE), PSO and DE-PSO, have been proposed by Xu and his colleagues (Xu et al., 2007a). This paper reports that hybrid DE-PSO training algorithm perform better with RNN for GRN inference. Another improvement in RNN training algorithm has been proposed by Ghazikhani and his colleagues (Ghazikhani et al., 2011). They proposed a multi-agent system trainer based on multi-population PSO algorithm. The result shows improvement on SOS repair network of e.coli. A more recent work by Noman and his colleagues (Noman et al., 2013) proposed a decoupled-RNN model of GRN. Here, decoupled means dividing the estimation problem of parameters for the complete network into several sub-problems, each of which estimate parameters associated with single gene. This decoupled approach decreases the dimensionality problem and makes the reconstruction of large network feasible. In our one of recent work, we applied evolutionary based approach, called ant colony optimization (ACO), for finding key interaction between genes (Raza et al., 2014). The result shows ACO-based algorithm is able to extract some key interactions between genes on real data. In this paper, we introduced gene extended Kalman filter for weight update in backpropagation through time (BPTT) training algorithm to improve the performance of RNN.

## 3. Materials and Methods

The Recurrent Neural Network (RNN) offers a good settlement between biological closeness and mathematical flexibility for depicting gene regulatory networks (GRNs) from gene expression data. For a network of non-linear processing elements (genes), RNN is able to capture the non-linear and dynamic interactions among genes (Noman et al., 2013). The RNN's architecture is potential to deal with temporal behavior. Using RNN for GRN model are mainly concerned with the capability of RNN to interpret complex temporal behavior (Hu et al., 2005). Since, GRN is non-linear and feedback in nature; hence RNN model is quite suitable for modeling GRNs (Ghazikhani et al., 2011). RNNs have recurrent connection which gives the ability to produce oscillatory and periodic activities. The complexity can be handled in the time domain and network behavior may be administered with the help of set of differential equations (Lee & Yang, 2008).

### 3.1 Recurrent Neural Networks

The architecture of RNN is different from the architecture of feedforward neural network. The RNN not only work on an input space but also on an internal state space. An internal state space is nothing but a trace of what has already been processed by network. Neurons in RNN can be connected to any other neuron in the same or a previous layer. RNN may be trained with a variant of backpropagation algorithm, called Backpropagation Through Time (BPTT).

The internal state space helps in representation and learning of temporally (or sequentially) extended dependencies over unspecified intervals according to,

$$y(t) = G(s(t)) \qquad (1)$$
$$s(t) = F(s(t-1), x(t)) \qquad (2)$$

where, $s(t)$ is internal state which is a function of previous state at time '$t–1$' as current input is $x(t)$.

In a simple recurrent network, input vector is similarly propagated through a weight layer, but in addition, it is also combined with previous state activation via an additional recurrent weight layer U, such as,

$$net_j(t) = \sum_i^n x_i(t)v_{ji} + \sum_h^m y_h(t-1)u_{jh} + \theta_j \qquad (3)$$

where $m$ is the number of 'state' nodes. A typical diagram of RNN is shown in Fig. 1.

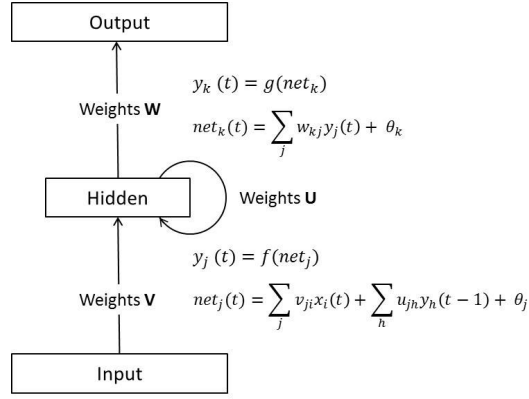

**Fig. 1** A recurrent neural network

The output of network can be computed by state and a set of output weights W, given by,

$$y_k(t) = g(net_k(t)) \qquad (4)$$

$$net_k(t) = \sum_{j}^{m} y_j(t) w_{kj} + \theta_k \qquad (5)$$

where $g$ is an output function (probability similar to $f$).

### 3.2 RNN based GRN model and its working

Using RNN as a GRN model is based on assumption that regulatory effect on the expression of a particular gene is represented in the form of a neural network where node represents gene and interconnection between them defines regulatory interaction. In RNN based model of GRN, expression of genes at time 't' may be computed from a gene node and the output of a gene node at time 't+Δt' may be calculated from expressions and connection weights (regulatory effects) of all the genes connected to a given gene at time 't'. Hence, regulatory effect to a particular gene is considered as weighted sum of all genes regulating that gene. Further, these regulatory effects are converted to a normalized value between 0 and 1 by using a sigmoid transfer function. For a system with continuous time, GRN models are represented as (Rui et al., 2004; Hu et al., 2005; Noman et al., 2013; Raza, 2014),

$$\tau_i \frac{de_i}{dt} = f\left[\sum_{j=1}^{n} w_{ij} e_j + \sum_{k=1}^{k} v_{ik} u_k + \beta_i\right] - \lambda_i e_i \qquad (6)$$

where, $e_i$ is expression level of gene $i$ ($1 \leq i \leq n$), $n$ is total number of genes, $f(.)$ is a non-linear transfer function, generally sigmoidal function $f(z) = 1/(1+e^{-z})$, $w_{ij}$ is synaptic weight representing the regulatory effect of $j^{th}$ gene on $i^{th}$ gene ($1 \leq i, j \leq n$), $u_k$ is $k^{th}$ ($1 \leq k \leq K$) external variable, $v_{ik}$ is effect of $k^{th}$ external variable on gene $i$, $\tau$ is time constant, $\beta$ is bias term and $\lambda$ is decay rate parameter (Rui et al., 2004).

A negative (or positive) value of synaptic weight $w_{ij}$ shows inhibition (or activation) of $j^{th}$ gene on $i^{th}$ gene. The $w_{ij}=0$ means that $j^{th}$ gene has no regulatory effect on $i^{th}$ gene. Because of limited time-point in expression data, the model equation (6) can be written in discrete form for simplicity and computation convenience as,

$$e_i(t + \Delta t) = \frac{\Delta t}{\tau_i} f\left[\sum_{j=1}^{n} w_{ij} e_j(t) + \sum_{k=1}^{k} v_{ik} u_k(t) + \beta_i\right] + \left(1 - \lambda_i \frac{\Delta t}{\tau_i}\right) e_i(t) \qquad (7)$$

The architecture diagram of the proposed model is shown in Fig. 2. The model takes microarray datasets (time-series as well as steady state) as input, preprocesses and normalize it and then the model is trained with these datasets. The proposed model gives a weight matrix as output which is interaction (regulation) matrix representing gene-gene interactions. The weight matrix is discretized to 0 or 1 and discretized regulation matrix is fed to the network visualizer. Once the network is inferred, it can be evaluated with the experimentally validated network (true network). The entire flow of data between different components of the model is shown in Fig. 2.

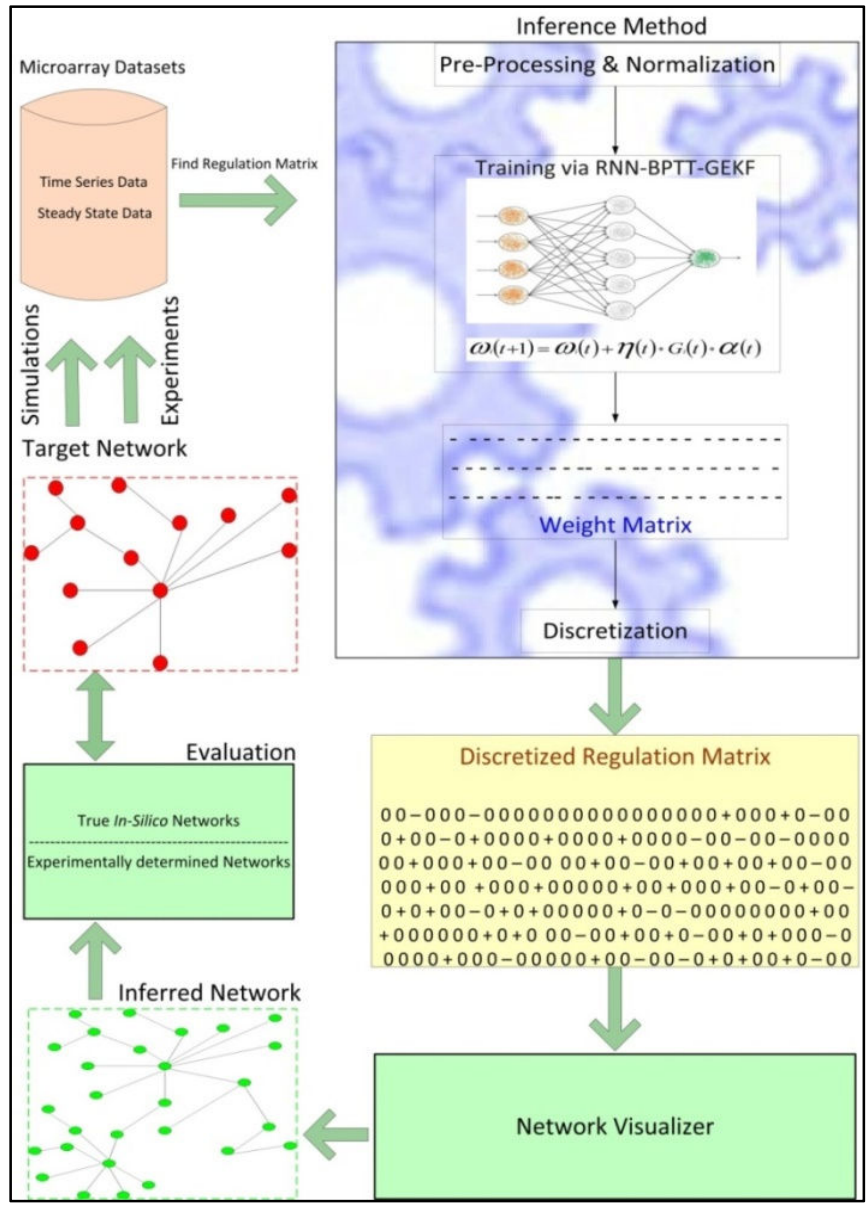

**Fig. 2** Proposed model architecture

### 3.3 Learning algorithm: Backpropagation Through Time

Once the network model is finalized, next step would be to set thresholds and time constants for every neuron in the network and weight of connections between neurons so as to achieve appropriate system behavior of the network. From the literature it has been observed that Backpropagation Through Time (BPTT) learning algorithm works well with RNN and are used for updating appropriate parameters of RNN in discrete-time steps. The BPTT is an extension of standard backpropagation algorithm and main concern behind BPTT is "unfolding" of discrete-time RNN (DTRNN) to a multilayer feed-forward network and by this way, the topology of the network increases by one layer at every time step. The objective of BPTT is to compute gradient over the trajectory and update connection weights accordingly. In BPTT, connection weights are updated once a forward step is completed. During these steps, activation is transmitted through network and every neuron stores its activation locally for entire length of trajectory. The connection weights can be updated in backward step by using following rules (Werbos, 1990):

$$\Delta w_{ij} = -\eta \frac{\partial E(t_0,t_1)}{\partial w_{i,j}} = \eta \sum_{\tau=t_0}^{t_1} \delta_j(\tau) x_i(\tau - 1) \qquad (8)$$

and,

$$\delta_j(\tau) = \begin{cases} f'(net_j(\tau))e_j(\tau) & if\ \tau = t_1 \\ f'(net_j(\tau))\left[e_j(\tau) + \sum_{l \in U}^{t_1} \delta_l(\tau+1)w_{i,l}\right] & if\ t_0 \leq \tau < t_1 \end{cases} \qquad (9)$$

where, $\delta_j$ stands for derivative of error surface, $\eta$ is learning rate, $e_j$ is corresponding error and $U$ is set of all outputs.

### 3.4 Kalman Filter

Kalman filter is a collection of mathematical equations which is known as one of the essential breakthrough in control theory, developed by E. Kalman in 1960. The initial applications of Kalman filter were in control of complex dynamical systems, such as manufacturing processes, aircrafts, ships or spaceships guidance and control. Recently, the extended Kalman filter has appeared in ANN learning algorithm. The Kalman filter comprises of a series of equations giving an effective and efficient recursive computational method to approximate the process state by minimizing mean squared error (MSE). This filter has been proved to be much powerful in many aspects: it can be used for estimation of present, past and future states (Welch & Bishop, 2001). The algorithm works in two-steps. First step is the "prediction" step that estimates current state variables, as well as their uncertainties. After the observation of next measurement, these estimates are updated by using a weighted average. Since, this algorithm works recursively, it needs only last "best guess" to estimate new state.

**3.4.1 The discrete Kalman filter algorithm:** The Kalman filter estimates a process by means of feedback control, i.e., estimation of process state at some given time and then finding a feedback in the form of noise measurements. The Kalman filter equations are clubbed into two groups: *time update* or "**predictor**" equations and *measurement update* or "**corrector**" equations. *Predictor* equations project forward (in time) current state estimate and error variance estimates to get *a priori* estimates for next time-step. *Corrector* equations are responsible for feedback – i.e., integrating a new measurement into *a priori* estimate to get an improved *a posteriori* estimate. The Kalman filter cycle of two major steps (e.g. predictor and corrector) with a complete set of operations in each step is shown in Fig. 3.

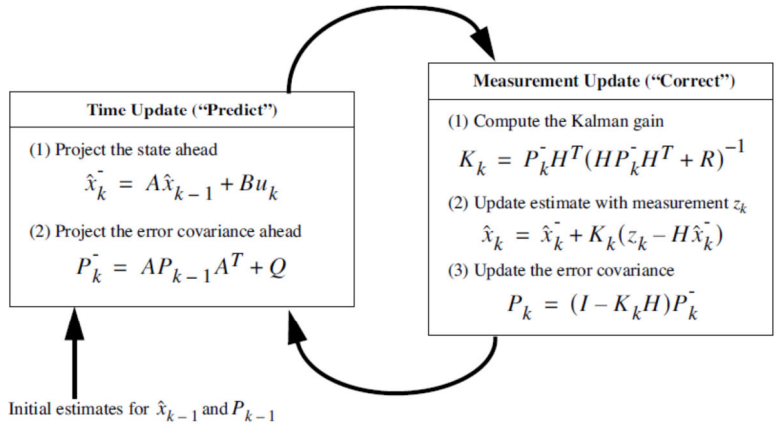

**Fig. 3** A complete view of operations in Kalman filter (Welch & Bishop, 2001)

**3.4.2 Generalized extended Kalman filter and weight update function for RNN:** The Kalman filter estimates optimally for linear system models only with some independent white noise. But in life sciences and engineering, mostly systems are nonlinear and hence simple Kalman filter fail. Few efforts were made by the researchers to apply Kalman filter to nonlinear systems and as a result generalized extended Kalman filter came to existence. The generalized extended Kalman filter (GEKF) is nonlinear form of Kalman filter that linearizes an estimate of current mean and covariance. In a well-defined transition models, GEKF is regarded as *de facto* standard in nonlinear state estimation theory. Due to better estimation even in noisy data and applicable to nonlinear systems, the GEKF can be used for weight update during RNN training. The weight update function for RNN training can be hybridized with the GEKF, given by,

$$w_i(t+1) = w_i(t) + G_i(t)\alpha(t)\gamma(t) \qquad (10)$$

*where*, $w_i(t)$ is weight vector at time '*t*' for i$^{th}$ gene, $G_i(t)$ is kalman gain, $\alpha(t)$ is invocations and $\gamma(t)$ is learning rate. The Kalman gain $G_i(t)$ is given by,

$$Kalman\ Gain,\ \ G_i(t) = K_i(t, t-1)C_i^T(t)\tau(t) \qquad (11)$$

where, $K_i$ is error covariance matrix of i$^{th}$ gene, $C(t)$ is *p*-by-*W* measurement matrix and $\tau(t)$ is learning rate. The invocations $\alpha(t)$ is given by,

$$\alpha(t) = d(t) - \hat{d}(t) \qquad (12)$$

where, d(t) is desired output and $\hat{d}(t)$ is approximate output.

**3.5 Implementation**

The proposed RNN-based hybridized model has been implemented in Matlab software environment. The Matlab program takes training time-series gene expression data in .csv (Comma Separated Value) formats and produces mean weight of ten simulations in .csv formats. All the experiments were executed on Intel® Core i7 quad-core processor of 2.2 GHz with 8 GB of RAM. Each experiment was executed 10 times to get a normalized and reliable weight matrix. The obtained normalized weight matrix (say, adjacency matrix of graph) has been discretized. The discretized adjacency matrix is loaded into Cytoscape (Shannon et al., 2003) software tool for visualization as gene regulatory network.

**4. Results and Discussions**

This proposed model has been tested on two real network as well as two simulated benchmark datasets of DREAM challenge which are described in the following section.

**4.1 Results on real data of SOS DNA repair networks**

Here, we considered gene expression data of SOS DNA repair network of *e.coli* that has around 30 genes regulated at transcription level. By using these experiments, expressions of eight major genes have been documented (Ronen et al., 2002). These genes are uvrD, lexA, umuD, recA, uvrA, uvrY, ruvA, polB. If DNA damage takes place, protein RecA gets activated and mediate LexA autocleavage by binding to single-stranded DNA molecules. Protein LexA is a master repressor which represses all genes when no damage takes place. Fall in LexA expression-level forces activation of SOS genes. When damage is repaired, expression level of RecA drops, that cause accumulation of LexA. LexA binds site in promotor region of these SOS genes, represses their expression and then cells restored to their original states. LexA and RecA are hubs genes in the SOS regulatory pathway.

The proposed method has been trained with SOS DNA repair network dataset and weight matrix is obtained (see Annexure A1 in supplement file). Further, the obtained weight matrix has been discretized using Inter-Quartile discretization (see Annexure A2 in supplement file). From the result it is observed that most of the regulatory interactions were correctly identified by proposed hybrid algorithm. Inhibition of *lexA* gene on all other genes has been correctly identified. Similarly, activation of *recA* gene on all other genes, except on polB, was identified successfully. In addition, prediction also has a number of false positives that may be either unknown regulations or due to effect of noises in the data. Out of 16 interactions, as reported in the literature, our method identified 15 interactions correctly. The performance of proposed method in terms of sensitivity, specificity, precision, recall and F-score is shown in Table 1. From Table 1, it can be observed that the sensitivity and recall both are 1.0 because total number of identified FNs is zero. Results also identified 16 FPs, due to that the value of specificity and precision are low.

**Table 1.** Performance of proposed method on SOS network

| Sensitivity TP/(TP+FN) | Specificity TN/(TN+FP) | Precision TP/(TP+FP) | Recall TP/(TP+FN) | F-Score 2*P*R/(P+R) |
|---|---|---|---|---|
| 1.0 | 0.67 | 0.48 | 1.0 | 0.65 |

We compared our results, in terms of truly identified regulations, with others work which are shown in Fig. 4. The comparison shows the superiority of our results over the others. The identified regulations in SOS network by different researchers has been shown in Table 2. Recent work by Noman et al., 2013 has predicted 9 regulations, i.e., inhibition of lexA on urvD, umuDC, recA, uvrY, ruvA and polB; and activation of recA on lexA, uvrY and polB.

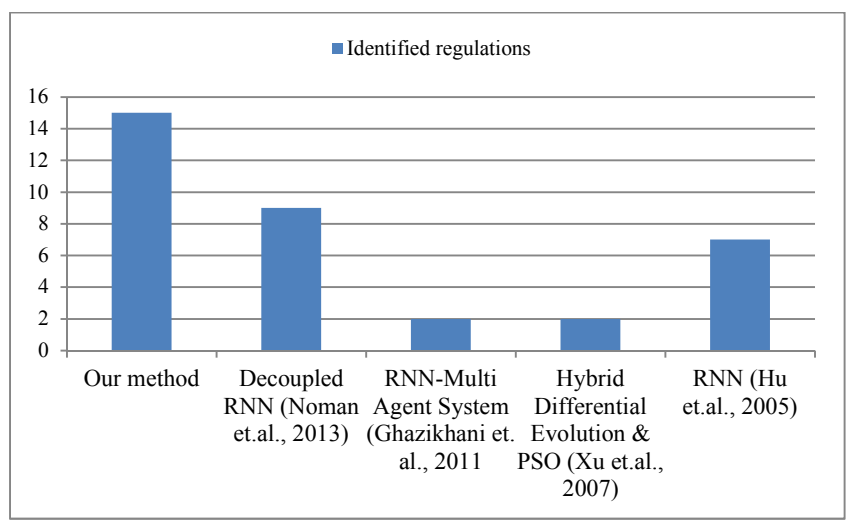

**Fig. 4** Comparison of the obtained results with others

**Table 2** Comparison of identified number of regulations in SOS network

| Methods | Descriptions |
|---|---|
| Our method | Predicts 15 regulations<br>- lexA inhibits all other genes in SOS network including self.<br>- recA activates all other genes including itself except polB. |
| Decoupled RNN<br>(Noman et al., 2013) | Predicts 9 regulations<br>- lexA inhibits uvrD, umuDC, recA, uvrY, ruvA and polB.<br>- recA activates lexA, uvrY and polB |
| RNN-Multi Agent System<br>(Ghazikhani et al., 2011 | Predicts 2 regulations<br>- lexA inhibits recA and polB |
| Hybrid Differential Evolution & PSO<br>(Xu et al., 2007) | Predicts 2 regulations<br>- lexA inhibits uvrD and<br>- recA activates lexA |
| RNN<br>(Hu et al., 2005) | Predicts 7 regulations<br>- lexA inhibits umuDC, ruvA and polB.<br>- recA activates lexA, umuDC, recA and uvrY |
| RNN-PSO-BPTT<br>(Xu et al., 2004) | Predicts 4 regulations PSO<br>- lexA inhibits uvrD, recA and ruvA<br>- recA activates lexA<br><br>Predicts 4 regulations BPTT<br>- lexA inhibits umuDC, uvrA and polB<br>- recA activates lexA |

Fig. 5 shows plot of real gene expression profiles versus learned mean expression profiles. It is observed that the method adequately captures the dynamics of all genes. The dynamics of gene urvD, umuDC, recA and polB are well learned with major changing trends of gene expression levels reflected in the learning curve. The expression profiles for gene ruvA oscillate between maximum value and zero.

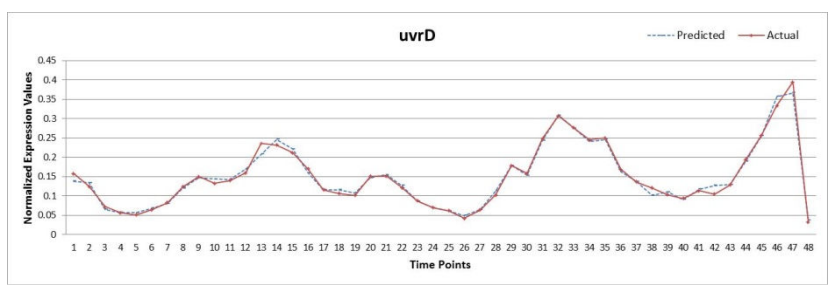

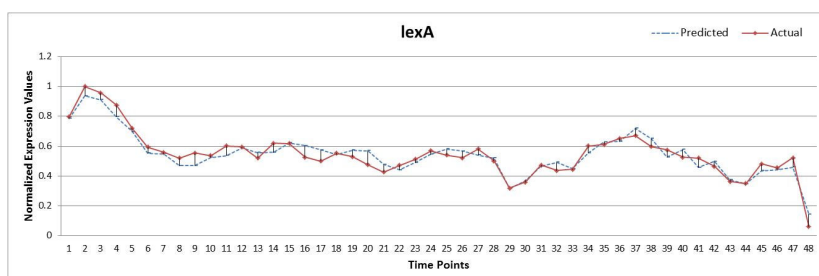

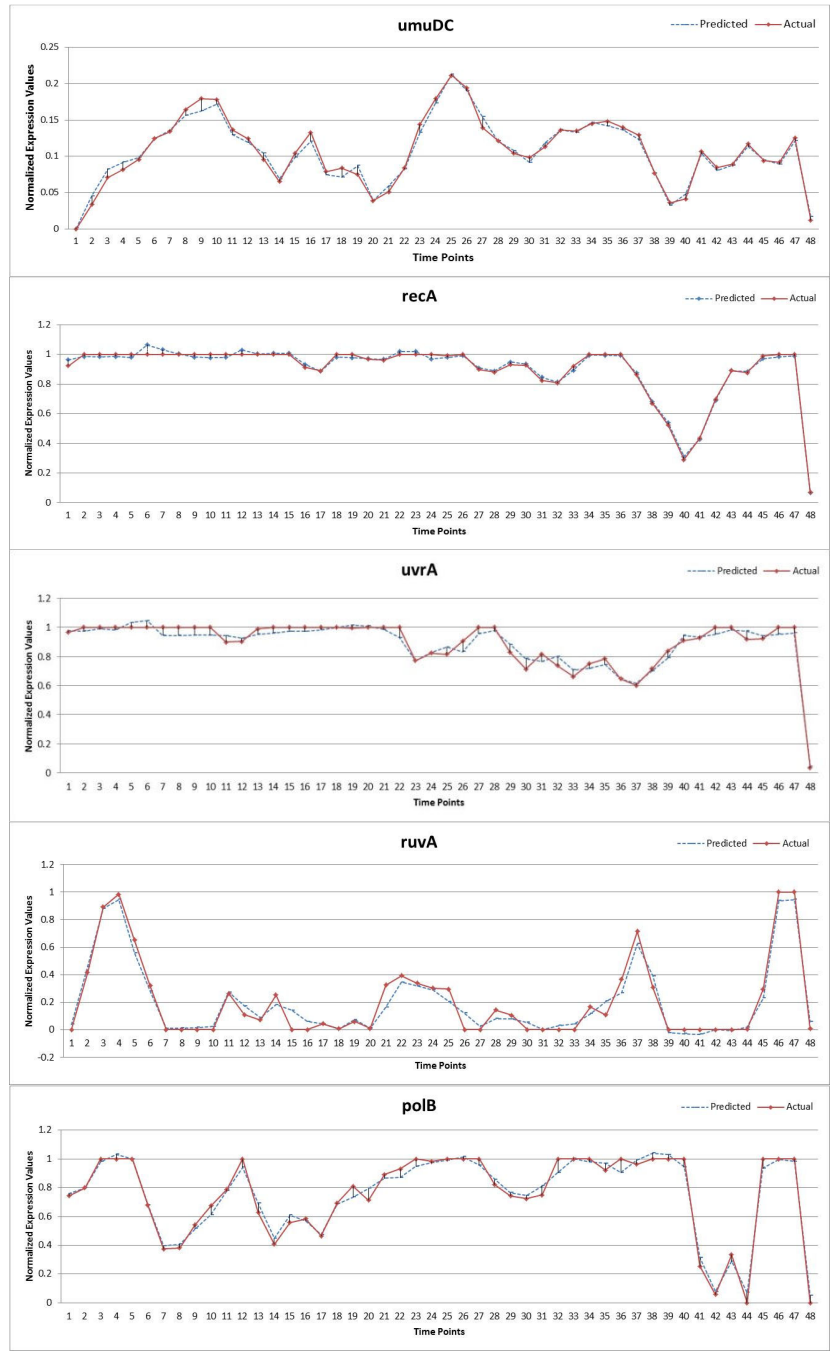

**Fig. 5** The comparison of dynamic behaviour in terms of gene expression level between learned network and the real GRN. Here x-axis represent samples at six minute internal and y-axis represents normalized expression values.

### 4.2 Results of in-vivo reverse-engineering and modeling assessment yeast network (IRMA)

We took IRMA network (Cantone et al., 2009) consists of five genes that regulate at least one another gene in the network. Expression level within the network is activated in the presence of glucose, allowing two time-series datasets to be generated. In first dataset, cells were grown-up in a glucose medium and switched to galactose (switch ON) with gene expression measured over 16 time-point using quantitative RT-PCR method. In next dataset, cells has been grown in a galactose medium and switched to glucose (switch OFF) with expression measured over 21 time points. These data has been widely used in the literature for the assessment of state-of-the-art GRN modeling and reverse-engineering methods.

The proposed model has been trained with both IRMA switch "OFF" and switch "ON" datasets and weight matrices are obtained. When the model is trained with IRMA switch "OFF" data, the identified weight matrix and their discretized values are obtained (see Annexure A3 and A4 in supplement file). Out of 8 interactions in the gold standard, the algorithm predicts 7 interactions (TPs) correctly with few false positives. We also compared the result on IRMA switch "OFF" network with the work of others, as shown in Table 3. Our result shows sensitivity of 0.88, specificity of 0.82, precision of 0.70 and F-score of 0.78, which is slightly better than the others.

**Table 3** Performance measures based on IMRA switch "OFF" data set with other state-of-the-art techniques

| Techniques | Original Network | | | |
|---|---|---|---|---|
| | Sensitivity | Specificity | Precision | F-Score |
| **Proposed RNN Hybrid Model** | **0.88** | **0.82** | **0.70** | **0.78** |
| (Morshed & Chetty, 2011 June) | 0.75 | 0.82 | 0.62 | 0.71 |
| BITGRN (Morshed & Chetty, 2011) | 0.63 | 0.71 | 0.50 | 0.56 |
| TDARACNE (Zoppoli et al., 2010) | 0.60 | - | 0.37 | 0.46 |
| NIR & TSNI (Gatta et al., 2008) | 0.38 | 0.88 | 0.60 | 0.47 |
| ARACNE (Margolin et al., 2006) | 0.33 | - | 0.25 | 0.28 |
| BANJO (Yu et al., 2004) | 0.38 | 0.88 | 0.60 | 0.46 |

The Fig. 6 shows actual versus predicted normalized gene expression values in case of switch "OFF" data. It can be observed that hybrid model adequately captures the dynamics of all five genes. The dynamics of genes CBF1, SWI5 and ASH1 are well learned and expression value of genes SWI5 and ASH1 gradually drop to zero.

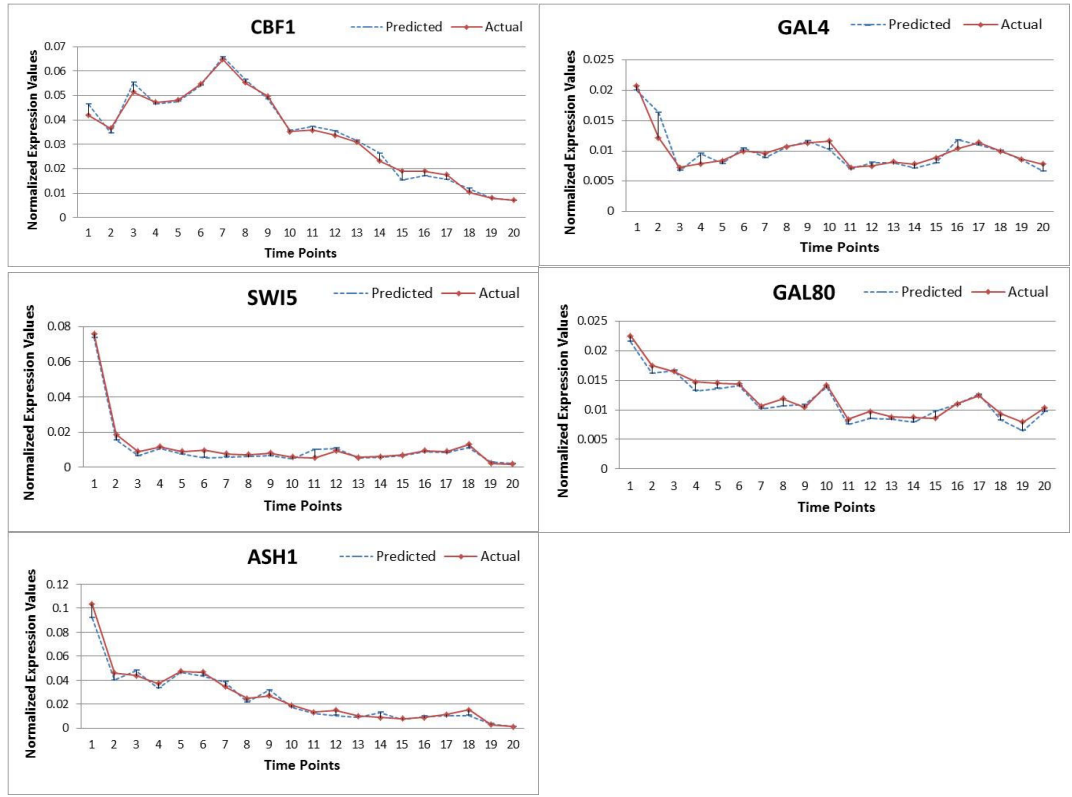

**Fig. 6** Comparison of dynamic behavior between learned network and actual network in IRMA network. Here, x-axis represents samples at 21 time-points and y-axis represents normalized expression value.

Similarly, we trained the proposed hybrid RNN model with IRMA switch "ON" data and weight matrix and their discretized values are identified (see Annexure A5 and A6 in supplement file). Out of 8 interactions, 7 has been correctly identified with 2 FPs and 1 FN. The performance of the algorithm on IRMA switch "ON" data in terms of sensitivity, specificity, precision and F-score is shown in Table 4. The proposed hybrid model achieved sensitivity of 0.88, specificity of 0.88, precision of 0.78 and F-score of 0.82. We also compared obtained result with other state-of-the-art techniques, as shown in Table 4. The result shows the superiority of proposed hybrid model over others.

**Table 4** Performance comparison of IMRA switch "ON" network with other state-of-the-art techniques

| Techniques | Original Network | | | |
|---|---|---|---|---|
| | Sensitivity | Specificity | Precision | F-Score |
| **Proposed RNN Hybrid Model** | **0.88** | **0.88** | **0.78** | **0.82** |
| Morshed & Chetty, 2011 June | 0.75 | 0.88 | 0.75 | 0.75 |
| BITGRN (Morshed & Chetty, 2011) | 0.63 | 0.94 | 0.83 | 0.71 |
| TDARACNE (Zoppoli et al., 2010) | 0.63 | 0.88 | 0.71 | 0.67 |
| NIR & TSNI (Della et al., 2008) | 0.50 | 0.94 | 0.80 | 0.62 |
| ARACNE (Margolin et al., 2006) | 0.60 | - | 0.50 | 0.54 |
| BANJO (Yu et al., 2004) | 0.25 | 0.76 | 0.33 | 0.27 |

Fig. 7 depicts actual versus predicted gene expression values in case of IRMA switch "ON" data. It is observed that our hybrid model truly captures the dynamics of all five genes. The expression of genes CBF1, SWI5 and ASH1 show increasing trends in expression values over time-points, while genes GAL5 and GAL80 show decreasing trends in expression.

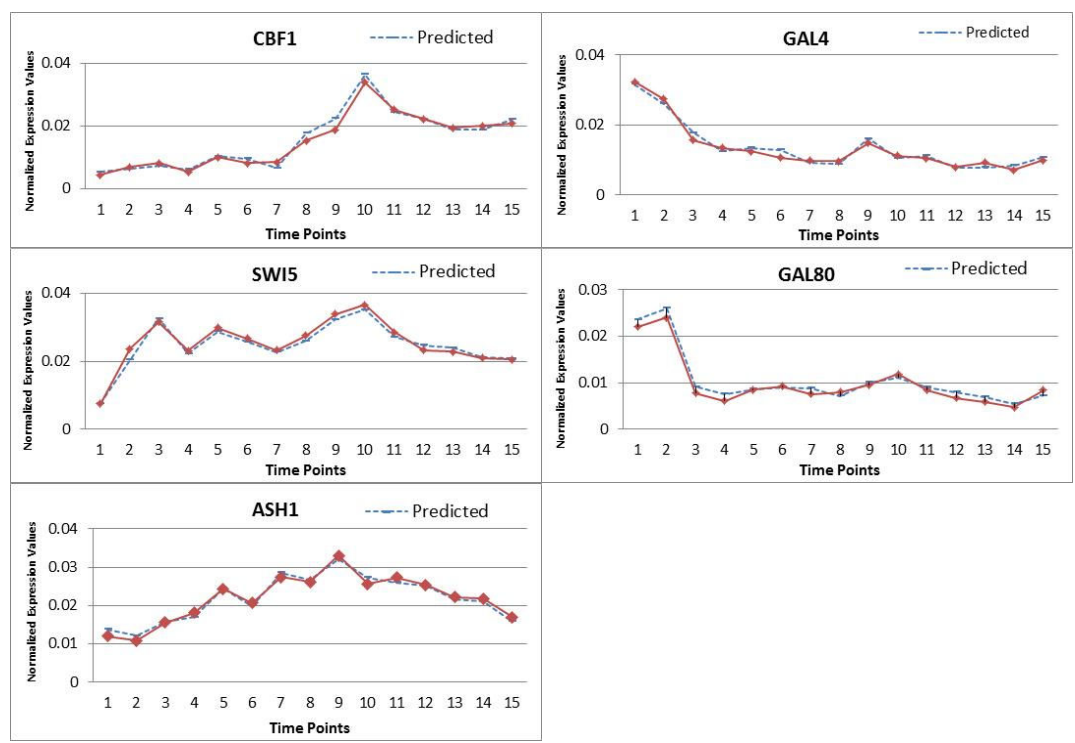

**Fig. 7** The comparison of dynamic behaviour between learned network and actual network in IRMA switch "ON" network. Here, x-axis represents samples at 15 time-points and y-axis represents normalized expression value.

### 4.3 Results on *in silico* networks

Here, we took two simulated networks: (i) small network consists of 10 genes (ii) mid-size network consists of 50 genes. These benchmark datasets has been published as part of DREAM4 challenge which is also downloadable from DREAM Project homepage (Stolovitzky et al., 2007). The datasets and their topologies have been taken from known GRNs of *e.coli* and *s.cerevisiac*. The time-series dataset consists of uniformly sample measurements (101 points, single replica) simulated using a parameterized set of SDEs in GeneNetworkWeaver (GNW) software tool (Schaffter et al., 2011).

*4.3.1 DREAM4 10-genes network:* The gold standard of this network consists of 10 genes and 15 interactions. After training the network, weight matrix is inferred and normalized (see Annexure A7 and A8 in supplement file). Out of 15 interactions, 14 were correctly identified with 13 FPs. The result shows sensitivity of 0.93 and specificity of 0.85, as shown in Fig. 8. However, the precision is only 0.52 because of large number of FPs.

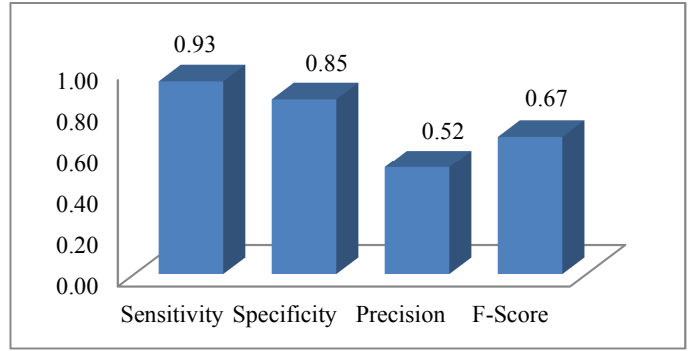

**Fig. 8** Performance measure of *in silico* network of 10 genes

*4.3.2 DREAM3 50-genes network:* The gold standard of this network contains 125 interactions. The proposed model infers 121 interactions (TPs) with large number of FPs, i.e., 662. The sensitivity and specificity goes down to 0.76 and 0.72, respectively. When results of the two *in silico* network (10-gene and 50-gene network) are compared, we found that as the size of the network grows, their complexity increases and hence prediction accuracy in terms of sensitivity, specificity, precision and F-score goes down.

### 4.4 Robustness of the model

To check the robustness of the proposed hybrid model, we added 5% Gaussian noise with standard deviation in 10-genes *in silico* data and then trained the model with the noisy data. The result shows that model is able to tolerate noises in the data. Out of 15 interactions, 14 were correctly predicted. From Fig. 9, we can see that both noise-free and noisy data achieved similar sensitivity of 0.93, while specificity, precision and F-score goes a bit down in noisy data. Hence, result shows that the proposed hybrid model is tolerant to noise and hence it is robust.

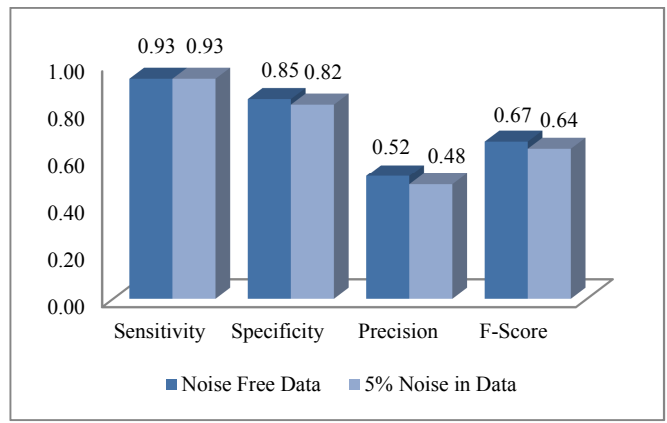

**Fig. 9** Performance measures in noisy-free and noisy data in 10 gene network

## 5. Conclusions

This paper proposes a recurrent neural network (RNN) based gene regulatory network (GRN) model hybridized with generalized extended Kalman filter for weight update in backpropagation through time training algorithm. The RNN is able to capture complex, non-linear and dynamic relationship among variables. Since, GRN is complex, relationship between genes are highly non-linear and dynamic in nature. Gene expression data are inherently noisy and Kalman filter performs well for estimation even in noisy data. Hence, non-linear version of Kalman filter, i.e., generalized extended Kalman filter has been applied for weight update during RNN training.

The proposed hybrid model has been tested on four benchmark datasets: i) DNA SOS repair network of *e.coli,* ii) IRMA network, iii) *in silico* network of 10-genes and iv) *in silico* network of 50-genes. Results show that model performs well on all four datasets. On the mid-size network of 50-genes, the predicted accuracy goes down in comparison to 10-gene network. To test the robustness of the model, we added 5% Gaussian noise in 10-gene network dataset and performance of the prediction is assessed. The result shows that added noise has negligible effect on the predicted results. From the results we observed that as the size of the network grows, the complexity also grows and prediction performance goes down. Modeling of large size GRN is still a challenge. The other problems are with microarray data such as i) curse of dimensionality problem; ii) data are inherently noisy due to experimental limitations and iii) data are not much reliable.

## References

Akutsu, T., Miyano S., & Kuhara, S. (1999). Identification of genetic networks from a small number of gene expression patterns under the Boolean network model. *Proc. of Pacific Symposium on Biocomputing*, 17-28.

Bower, J.M., & Bolouri, H. (2004). Computational modeling of genetic and biochemical networks. MIT Press, London.

Cantone, I., et al. (2009). A yeast synthetic network for in vivo assessment of reverse-engineering and modeling approaches. *Cell*, 137(1), 172-181.

Chen, T., He, H.L., & Churck, G.M. (1999). Modeling gene expression with differential equations. *Proc. of Pacific Symposium on Biocomputing,* 29-40.

Chiang, J-H., Chao, S-Y. (2007). Modeling human cancer-related regulatory modules by GA-RNN hybrid algorithms. *BMC Bioinformatics,* 8:91.

Cho, K-H., Choo, S-M., et.al. (2007). Reverse engineering of gene regulatory networks. *IET Systems Biology,* 1(3), 149–163.

Datta, D., et. Al. (2009). A recurrent fuzzy neural model of a gene regulatory network for knowledge extraction using differential equation. *Proc. of IEEE Congress on Evolutionary Computation,* 2900-2906.

Friedman, N., Linial, M. & Nachman, I.P. (2000). Using Bayesian networks to analyze expression data. *Journal of Computational Biology,* 7, 601-620.

Ghazikhani, A., Akbarzadeh Totonchi, M. R., & Monsefi, R. (2011). Genetic Regulatory Network Inference using Recurrent Neural Networks trained by a Multi Agent System. *Proc. of 1st International eConference on Computer and Knowledge Engineering (ICCKE) 2011*, 95-99.

Hu, X., Maglia, A., & Wunsch, D.A. (2005). A general recurrent neural network approach to model genetic regulatory networks. *Proc. of IEEE Conference on Engineering in Medicine and Biology 27$^{th}$ Annual Conference*, 4735-4738.

Huang, J., Shimizu, H., & Shioya, S. (2003). Clustering gene expression pattern and extracting relationship in gene network based on artificial neural networks. *Journal of Bioscience and Bioengineering,* 96, 421-428.

Husmeier, D. (2003). Sensitivity and specificity of inferring genetic regulatory interactions from microarray experiments with dynamic Bayesian networks. *Bioinformatics,* 19, 2271-2282.

Jong, H. &Page, M. (2008). Search for steady states of piecewise-linear differential equation models of genetic regulatory networks, *IEEE/ACM Transaction on Computational Biology and Bioinformatics,* 5(2), 208-222.

Jong, H. (2002). Modeling and simulation of genetic regulatory systems: A literature review. *Journal of Computational Biology*, 9:67-103.

Jung S.H. & Cho, K-H. (2007). Reconstruction of gene regulatory networks by neuro-fuzzy inference system. *Frontiers in the Convergence of Bioscience and Information Technologies,* 32-37.

Karlebach, G. & Shamir, R. (2008). Modeling and analysis of gene regulatory networks. *Nature Reviews Molecular Cell Biology,* 9, 770-780.

Kauffman, S.A. (1969). Metabolic stability and epigenesis in randomly constructed genetic nets. *Journal of Theoretical Biology*, 22, 437-467.

Keedwell, Ed., Narayanan, A. & Savic, D. (2002). Modeling gene regulatory data using artificial neural networks. *IEEE/INNS/ENNS International Joint Conference on Neural Networks* (IJCNN'02), 183-189.

Kentzoglanakis, K. (2012). A swarm intelligence framework for reconstructing gene networks: searching for biologically plausible architectures. *IEEE/ACM Transactions on Computational Biology and Bioinformatics,* 9(2), 358-371.

Kim, S., et. al. (2000). Multivariate measurement of gene expression relationships. *Genomics,* 67, 201-209.

Koch, I., Schueler, M. & Heiner, M. (2005). STEPP – search tool for exploration of Petri net paths: a new tool for Petri net-based path analysis in biochemical networks. *In Silico Biology*, 5, 129-137.

Lee, W-P & Yang, K-C. (2008). A clustering-based approach for inferring recurrent neural networks as gene regulatory networks. *Neurocomputing*, 71, 600-610.

Liang, S., Fuhrman, S. & Somogyi, R. (1998). REVEAL, a general reverse engineering algorithm for inference of genetic regulatory network architectures. *Pacific symposium on biocomputing*, World Scientific Publishing 3:18-29.

Liu, G., et. al. (2011). Combination of neuro-fuzzy network models with biological knowledge for reconstructing gene regulatory networks. *Journal of Bionic Engineering,* 8(1), 98-106.

Maraziotis I.A., Dragomir, A., & Thanos, D. (2010). Gene regulatory networks modeling using a dynamic evolutionary hybrid. *BMC Bioinformatics,* 11:140.

Martin S, Shang Z, Martino A, Faulon J-L (2007) Boolean dynamics of genetic regulatory networks inferred from microarray time series data. *Bioinformatics,* 23, 866-874.

Mitra, S., Das, R., & Hayashi, Y. (2011). Genetic networks and soft computing. *IEEE/ACM Transaction on Computational Biology and Bioinformatics*, 8(1):94-107.

Noman, N, Palafox, L, Iba, H. (2013). Reconstruction of gene regulatory networks from gene expression data using decoupled recurrent neural network model. In *Natural Computing and Beyond*, Proceedings in Information and Communications Technology, Springer, 6, 93–103.

Raza, K. (July 2014). *Soft Computing approach for modeling biological networks*. PhD Thesis, Jamia Millia Islamia, New Delhi, India.

Raza, K., & Kohli, M. (2014). Ant colony optimization for inferring key gene interactions. *arXiv Preprint arXiv*:1406.1626.

Raza, K., & Parveen, R. (2012). Evolutionary algorithms in genetic regulatory networks model. *Journal of Advanced Bioinformatics Applications and Research*, 3(1), 271-280.

Raza, K., & Parveen, R. (2013). Soft computing approach for modeling genetic regulatory networks. In *Advances in Computing and Information Technology*, 178, 1-11, Springer Berlin Heidelberg.

Remy, E., et al. (2006). From logical regulatory graphs to standard Petri nets: Dynamical roles and functionality of feedback circuits. Springer LNCS 4230:56-72.

Ronen, M., Rosenberg, R., Shraiman, B. & Alon, U. (2002). Assigning numbers to the arrows: Parameterizing a gene regulation network by using accurate expression kinetics. *Proc. Nat. Acad. Sci. USA*, 99(16), 10555–10560.

Rui, X., Xiao, H., Wunsch, D.C. (2004). Inference of genetic regulatory networks with recurrent neural network models. *Proc. of 26th Annual International IEEE Conference on Engineering in Medicine and Biology Society, 2004. IEMBS '04.*, 2, 2905-2908.

Schaffter, T, Marbach, D, and Floreano, D. (2011). GeneNetWeaver: In silico benchmark generation and performance profiling of network inference methods. *Bioinformatics*, 27(16):2263-70.

Schlitt, T. & Brazma, A. (2007). Current approaches to gene regulatory network modeling. *BMC Bioinformatics,* 8(Suppl 6):S9.

Shannon, P., et.al. (2003). Cytoscape: a software environment for integrated models of biomolecular interaction networks. *Genome Research.* 13(11):2498-504.

Shmulevich I, Dougherty E R, Kim S, Zhang W (2002) Probabilistic Boolean networks: A rule-based uncertainty model for gene regulatory networks. *Bioinformatics,* 18:261-274.

Shmulevich, I. and Dougherty, E.R. (2010). Probabilistic Boolean networks: the modeling and control of gene regulatory networks, *SIAM 2010*.

Silvescu, A., & Honavar, V. (2001). Temporal Boolean network models of genetic networks and their inference from gene expression time series. *Complex System*, 13, 54–75.

Sirbu, A., Ruskin, HJ., & Crane, M. (2010) Comparison of evolutionary algorithms in genetic regulatory network model. *BMC Bioinformatics*, 11:59.

Stolovitzky G, Monroe D, Califano A. (2007). Dialogue on Reverse-Engineering Assessment and Methods: The DREAM of High-Throughput Pathway Inference", in Stolovitzky G and Califano A, Eds, *Annals of the New York Academy of Sciences*, 1115:11-22.

Swain, M. T., Mandel, J. J., & Dubitzky, W. (2010). Comparative study of three commonly used continuous deterministic methods for modeling gene regulation networks. *BMC Bioinformatics*, 11(459), 1–25.

Tian T, Burrage K (2003) Stochastic neural network models for gene regulatory networks. *Proc. of IEEE Congress on Evolutionary Computation,* 162-169.

Tyson J J, Csikasz-Nagy A, Novak B (2002) The dynamics of cell cycle regulation. *Bioessays,* 24(12):1095-1109.

Vohradsky, J. (2001). Neural network model of gene expression. *FASEB J.* 15, 846-854.

Weaver, DC, Workman, CT, & Stormo, GD. (1999). Modeling regulatory networks with weight matrices. *Proc. of Proc. Pacific Symp. Biocomputing,* 112-123.

Wei, G., Liu, D., & Liang, C. (2004). Charting gene regulatory networks: strategies, challenges and perspectives. *Biochemistry Journal*, 381:1–12.

Welch, G., & Bishop, G. (2001). An introduction to the Kalman filter. University of North Carolina, Department of Computer Science, Technical Report TR 95-041.

Werbos, P. J. (1990). Backpropagation through time: what it does and how to do it. *Proc. of the IEEE*, 78(10), 1550-1560.

Xu, R., Hu, X., & Wunsch, D. C. (2004). Inference of genetic regulatory networks with recurrent neural network models. *Proc. of 26th IEEE Annual International Conference on Engineering in Medicine and Biology Society 2004*, 2, 2905-2908.

Xu, R., Venayagamoorthy, G.K. & Wunsch, D.C. (2007a). Modeling of gene regulatory networks with hybrid differential evolution and particle swarm optimization. *Neural Networks*, 20, 917-927.

Xu, R.., Wunsch, D.C., & Frank R.L. (2007b). Inference of genetic regulatory networks with recurrent neural network models using particle swarm optimization. *IEEE/ACM Transactions on Computational Biology and Bioinformatics,* 4(4): 681-692.

Yu, J., Smith, V., Wang, P., Hartemink, A. and Jarvis, E. (2004). Advances to Bayesian network inference for generating causal networks from observational biological data. *Bioinformatics*, 20(18):3594.

Yun, Z., & Keong, K. C. (2004). Reconstructing Boolean Networks from Noisy Gene Expression Data Theory. In *8th International Conference on Control, Automation, Robotics and Vision Kunming*, China, 6–9.

Zhang, Y., et al. (2009). Reverse engineering module networks by PSO-RNN hybrid modeling. *BMC Genomics,* 10(Suppl 1):S15.

Zhou, X., et. al. (2004). A Bayesian connectivity-based approach to constructing probabilistic gene regulatory networks. *Bioinformatics* 20(17), 2918-292.

Zoppoli, P., Morganella, S. & Ceccarelli, M. (2010). TimeDelay-ARACNE: Reverse engineering of gene networks from time-course data by an information theoretic approach. *BMC Bioinformatics*, 11(1): 154.